\documentclass{llncs}
\usepackage{graphicx}
\usepackage{amsmath}
\usepackage{multirow}
\usepackage{booktabs}

\begin{document}

\title{An Improved Ensemble-Based Machine Learning Model with Feature Optimization for Early Diabetes Prediction}




\author{
Md.~Najmul~Islam\thanks{Corresponding author: cse\_182210012101019@lus.ac.bd} \and
Md.~Miner~Hossain~Rimon \and
Shah~Sadek--E--Akbor~Shamim \and
Zarif~Mohaimen~Fahad \and
Md.~Jehadul~Islam~Mony \and
Md.~Jalal~Uddin~Chowdhury
}

\institute{
Department of Computer Science and Engineering, Leading University, Sylhet, Bangladesh\\
\email{
cse\_182210012101019@lus.ac.bd,
cse\_182210012101033@lus.ac.bd,
cse\_182210012101006@lus.ac.bd,
zarifmohaimen2000@gmail.com,\\
mony\_cse@lus.ac.bd,
jalal\_cse@lus.ac.bd
}
}

\maketitle
\begin{abstract}
Diabetes is a serious worldwide health issue, and successful intervention depends on early detection. However, overlapping risk factors and data asymmetry make prediction difficult. To use extensive health survey data to create a machine learning framework for diabetes classification that is both accurate and comprehensible, to produce results that will aid in clinical decision-making. Using the BRFSS dataset, we assessed a number of supervised learning techniques. SMOTE and Tomek Links were used to correct class imbalance. To improve prediction performance, both individual models and ensemble techniques such as stacking were investigated. The 2015 BRFSS dataset, which includes roughly 253,680 records with 22 numerical features, is used in this study. Strong ROC-AUC performance of approximately 0.96 was attained by the individual models Random Forest, XGBoost, CatBoost, and LightGBM.The stacking ensemble with XGBoost and KNN yielded the best overall results with 94.82\% accuracy, ROC-AUC of 0.989, and PR-AUC of 0.991, indicating a favourable balance between recall and precision. In our study, we proposed and developed a React Native-based application with a Python Flask backend to support early diabetes prediction, providing users with an accessible and efficient health monitoring tool. 
\end{abstract}

\section{Introduction}
Diabetes is a group of interconnected disorders that are characterized by high blood sugar. Diabetes occurs when the pancreas is unable to produce sufficient insulin or is unable to produce insulin at all, or the body cells fail to respond appropriately to insulin \cite{1}. 
There are 2 type of diabetes Type 1 diabetes and Type 2 diabetes. Type 1 diabetes develops when the cells in the pancreas that secrete insulin are attacked and destroyed by the immune system \cite{2}. Type 2 diabetes, on the contrary, develops when the body fails to respond to insulin (insulin-saturated fat and salt, age, chronic stress, high blood pressure, and low-density lipoprotein (LDL, ”bad”) cholesterol levels). Type 1 diabetes is more likely to be preceded by more environmental causes and infections in children than type 2 diabetes.A report by the World Health Organization (WHO), published in 2024, quotes that. In the year 2022, 830 million people were diagnosed with diabetes, which was 200 million in 1990 \cite{2}. Fifty percent of the people suffering from diabetes are not being treated, which can lead to serious complications even now, such as blindness, kidney failure, heart attack, stroke, and amputation \cite{3}.

Diabetes is an emerging global health burden that needs early and accurate detection to reduce severe complications and improve outcomes \cite{4}. Though numerous research efforts have applied machine learning to predict diabetes, many of these studies are based on limited or outdated datasets that cannot capture the complexity and diversity of real-world populations \cite{5}. Such datasets are imbalance, noisy data, and overlapping health indicators, which reduce model reliability and generalization.

In this regard, the paper develops an optimized machine learning framework for the prediction of diabetes, using the BRFSS 2015 dataset with more than 250,000 . It aims at overcoming data imbalance and redundancy by advanced preprocessing, enhancing the prediction performance by using ensemble learning methods such as stacking, which could improve the accuracy, interpretability, and clinical utility of diabetes risk prediction models. Diabetes prediction using machine learning has also been done with the PIMA Indian and Vanderbilt datasets. The works follow the general pipeline of data preprocessing, feature selection, and classification using models like Logistic Regression, Random Forest, SVM, and ensembling techniques such as XGBoost and LightGBM with very high accuracy. Most of the models in existence are based on small or region-specific datasets, which do not allow for generalization into larger populations. Besides, challenges like data imbalance, noise, and low interpretability decrease the reliability of predictions. To overcome these gaps, this work uses the BRFSS 2015 dataset, applying SMOTE + Tomek Links and ensemble stacking techniques for building a more accurate, balanced, and interpretable diabetes prediction framework.
The main aim of this study is to develop an accurate and efficient diabetes prediction model using advanced machine learning and ensemble techniques. In the proposed system, major improvements will be made at the data preprocessing stage, in feature selection, and in class balancing to enhance predictive performance. The main contributions of this work are listed as below:

\begin{enumerate}
\item Applied an advanced preprocessing pipeline that removes duplicates, imputes missing values, and normalizes data in order to prepare clean and reliable.

\item Applied SMOTE + Tomek Links for effective handling of data imbalance and noise reduction.

\item Applying the Ensemble Feature Selection Technique to rank the most important diabetes predictors among all the variables: MI, RFE, and LASSO.

\item A stacking ensemble model is developed where KNN was combined with XGBoost and was the best performer. The accuracy score it returns is high, 94.8\%, with a ROC-AUC score of 0.989.

\item Built an interactive web application with a conversational chatbot for personalized health assistance and a prediction system where users input clinical factors to receive instant diabetes risk assessments. The platform provides doctors with feature importance visualizations to identify key risk factors, bridging the gap between advanced machine learning and practical clinical decision-making.
\end{enumerate}

\section{Methodology}
This section explains the overall process used in our study, including data collection, preprocessing, feature selection, model development, and evaluation. Each of the steps was designed carefully, ensure reliable and accurate results for diabetes classification using machine learning
\subsection{Data Source and Data Preprocessing}



The study used the BRFSS dataset, we assessed a number of supervised learning techniques. Class imbalance was addressed using SMOTE and Tomek Links were used to correct class imbalance. This is used for undersampling the majority of cases. this method normalized the feature and handled missing values correctly.
The 2015 BRFSS dataset which includes roughly 253,680 records with 22 numerical features.

\begin{figure}[h!]
\centering
\includegraphics[width=0.6\textwidth]{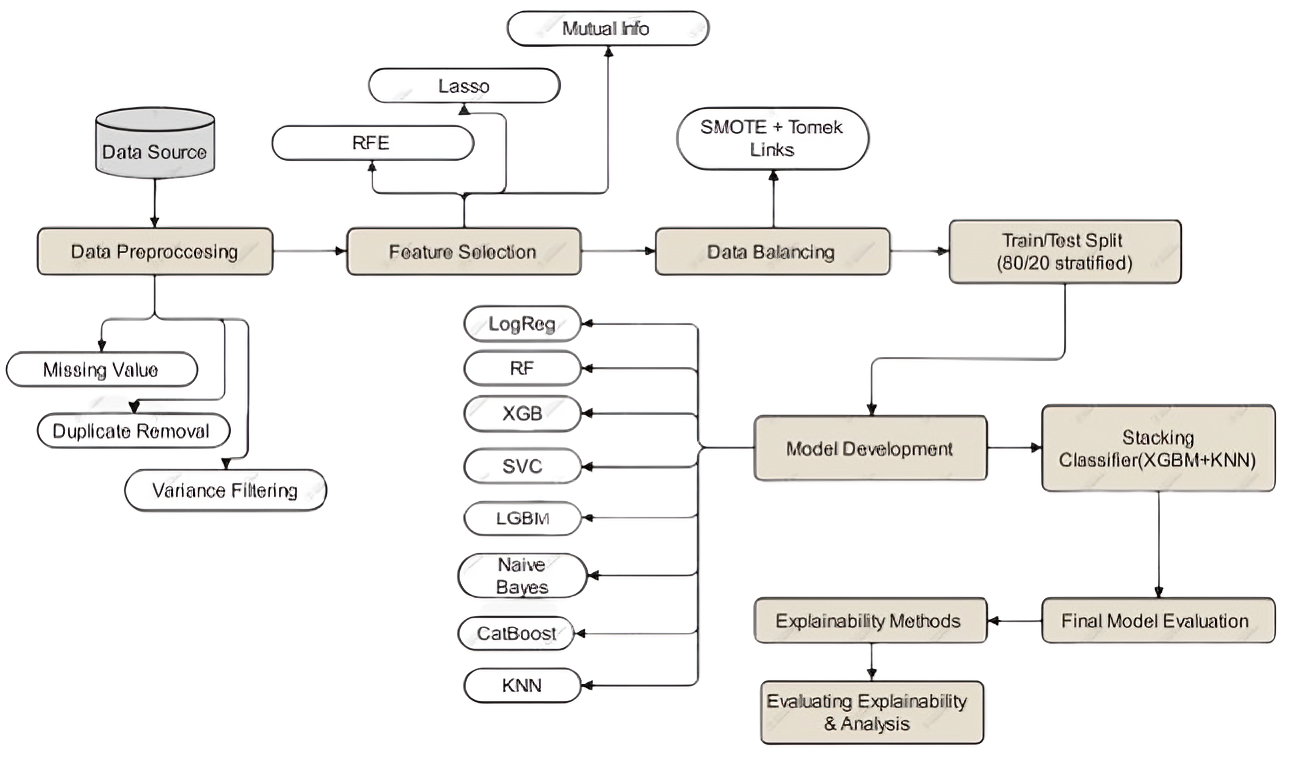}
\caption{Overall Workflow of Our Proposed Model}
\label{fig:workflow}
\end{figure}

\subsection{Analysis of Exploratory Data (EDA)}
Prior to modelling, we used EDA to understand data distributions, patterns, and quality. Univariate analysis. Distributions, imbalances, and outliers were observed using histograms for both continuous and categorical variables. Correlation analysis.  Correlation analysis and VIF identified multicollinearity, with some variables seemed less significant, others, like blood pressure, cholesterol, and BMI, revealed higher relationships with diabetes.

\subsection{Feature Selection}


We used three feature selection techniques to increase efficiency and decrease dimensionality: LASSO regression, Recursive Feature Elimination (RFE), and Mutual Information (MI). A single table was created by normalising the ranks derived from different techniques.

The top 18 predictors were chosen for model training based on the average ranking: ['GenHlth', 'HighBP', 'BMI', 'Age', 'HighChol', 'CholCheck', 'Income', 'Sex', 'HeartDiseaseorAttack', 'HvyAlcoholConsump', 'AnyHealthcare', 'DiffWalk', 'PhysActivity', 'Smoker', 'Veggies', 'Fruits', 'Education', 'Stroke']. This selection lowers the chance of overfitting while striking a balance between explainability and predictive power.

\subsection{Data Splitting}

About 80\% of the information was used for training and 20\% was used for testing. The test data was set aside for the final assessment to ensure that the findings were unbiased, while the training data was used to construct models and adjust hyperparameters using k-fold cross-validation.


\subsection{Model Training Strategy}



For our study, we applied several supervised machine learning algorithms for model development. These include logistic regression, random forest, Gradient Boosting, XGBoost, Support Vector Classifier (SVC), LightGBM, CatBoost, Naive Bayes, and K Nearest Neighbors (KNN). The aim was to compare both simple explainable models and advanced ensemble learners to achieve the best outcomes. 

For hyperparameter optimisation, we follow a two-step process. In the first step, we apply Randomized Search to broadly explore a wide range of hyperparameter space. GridSearchCV is used to fine-tune the models for optimal performance once the promising parameter has been identified. This approach ensures both computational efficiency and achieves the best performance. To prevent overfitting and ensure robustness, we use 5-fold cross-validation during training. The final model was evaluated on the unseen test sets, where multiple performance metrics were considered, including the classification report, confusion matrix, and ROC-AUC, to provide a fair and balanced assessment of the models.

\subsection{Development of the Stacking Ensemble Model}


We chose to use stacking instead of depending on a single model. This approach combines multiple algorithms by capturing their individual strengths. Looking at Table 1, XGBoost caught our attention. It provides balanced and solid performance across accuracy, precision, and F1-score. However, it had a weakness, recall was slightly low, but in medical applications, recall plays a very significant role. Even missing a single diabetes case may cause a problem. That's why we need to identify all of the positive cases. KNN took a different approach. It emphasizes recall, which means it holds diabetes patients really well. Overall, the weakness is slightly lower accuracy. We realized that two of these models will fulfill each other perfectly, so we stacked them together using LightGBM as a meta learner.

This combination worked well. By mixing XGBoost's balanced performance with KNN's strong recall, we built a more robust model. It holds more positive cases with maintains a solid accuracy.

\section{Results Analysis}
In this section, we will illustrate and discuss about the outcomes of our experiments. including setup, model evaluation, contribution, and comparison with the previous works. The goal is to analyze how each model performed, explain the key findings from the results. and highlight how our proposed stacking ensemble model overperformed for diabetes prediction. 

\subsection{Performance Analysis of All Models}


In this study, we used nine different types of machine learning models to predict diabetes based on features. We used supervised machine learning algorithms, including  Logistic Regression and Naive Bayes, which performed the lowest. Tree-based algorithms such as Random Forest, XGBoost, LightGBM, and CatBoost performed much better, with around  90-91\% accuracy. Among them, XGBoost provided the best precision, and KNN provided the highest recall. 
Finally, stacking ensemble model with XGBoost and KNN, which achieved the best outcome. This model achieved the highest accuracy of 94.82\% and also gave the highest F1-score and ROC-AUC. The stacking method worked well because XGBoost is used for accuracy, and KNN is used for strong recall. Overall, the ensemble model outperformed any single model.

\begin{table}[ht]
\centering
\caption{Performance Metrics of Applied Models for Early Diabetes Prediction}
\begin{tabular}{lccccc}
\toprule
Model & Accuracy & Precision & Recall & F1 & ROC-AUC \\
\midrule
LightGBM & 0.9108 & 0.9694 & 0.8483 & 0.9048 & 0.9651 \\
Random Forest & 0.9096 & 0.9389 & 0.8762 & 0.9064 & 0.9678 \\
CatBoost & 0.9074 & 0.9596 & 0.8506 & 0.9018 & 0.9619 \\
Logistic Regression & 0.7386 & 0.7257 & 0.7672 & 0.7459 & 0.8135 \\
Naive Bayes & 0.7129 & 0.6683 & 0.8453 & 0.7464 & 0.7750 \\
XGBoost & 0.9063 & 0.9642 & 0.8440 & 0.9001 & 0.9628 \\
KNN & 0.9023 & 0.8635 & 0.9557 & 0.9072 & 0.9417 \\
Stacking & \textbf{0.9482} & 0.9293 & \textbf{0.9702} & \textbf{0.9493} & \textbf{0.9895} \\
\bottomrule
\end{tabular}
\label{tab:performance}
\end{table}


Table~\ref{tab:performance}, demonstrates the performance of all nine models, where tree-based models performed much better. In addition, ensemble methods significantly outperformed traditional models.

\begin{table}[ht]
\centering
\caption{Training and Testing Accuracy Comparison of Applied Models}
\begin{tabular}{lcc}
\toprule
\textbf{Model} & \textbf{Train Accuracy} & \textbf{Test Accuracy} \\
\midrule
Stacking & \textbf{0.9875} & \textbf{0.9482} \\
LightGBM & 0.9185 & 0.9100 \\
Random Forest & 0.9584 & 0.9095 \\
CatBoost & 0.9396 & 0.9072 \\
XGBoost & 0.9203 & 0.9056 \\
KNN & 0.9919 & 0.8996 \\
Logistic Regression & 0.7413 & 0.7399 \\
Linear SVC & 0.7398 & 0.7369 \\
Naive Bayes & 0.7144 & 0.7129 \\
\bottomrule
\end{tabular}
\label{tab:train_test_accuracy}
\end{table}


Table~\ref{tab:train_test_accuracy}, demonstrated a significant point. It showed in the stacking model that the difference between training and testing accuracy is low, which is why it performed well on new data. But KNN provided the best performance in training, on the other hand, in testing, it performed poorly. So KNN showed overfitting individually. By contrast, the ensemble model did not show this problem and was more stable compared to traditional statistical approaches.

\subsection{Performance Measure for Ensemble Model}

\begin{figure}[h!]
\centering
\includegraphics[width=0.4\textwidth]{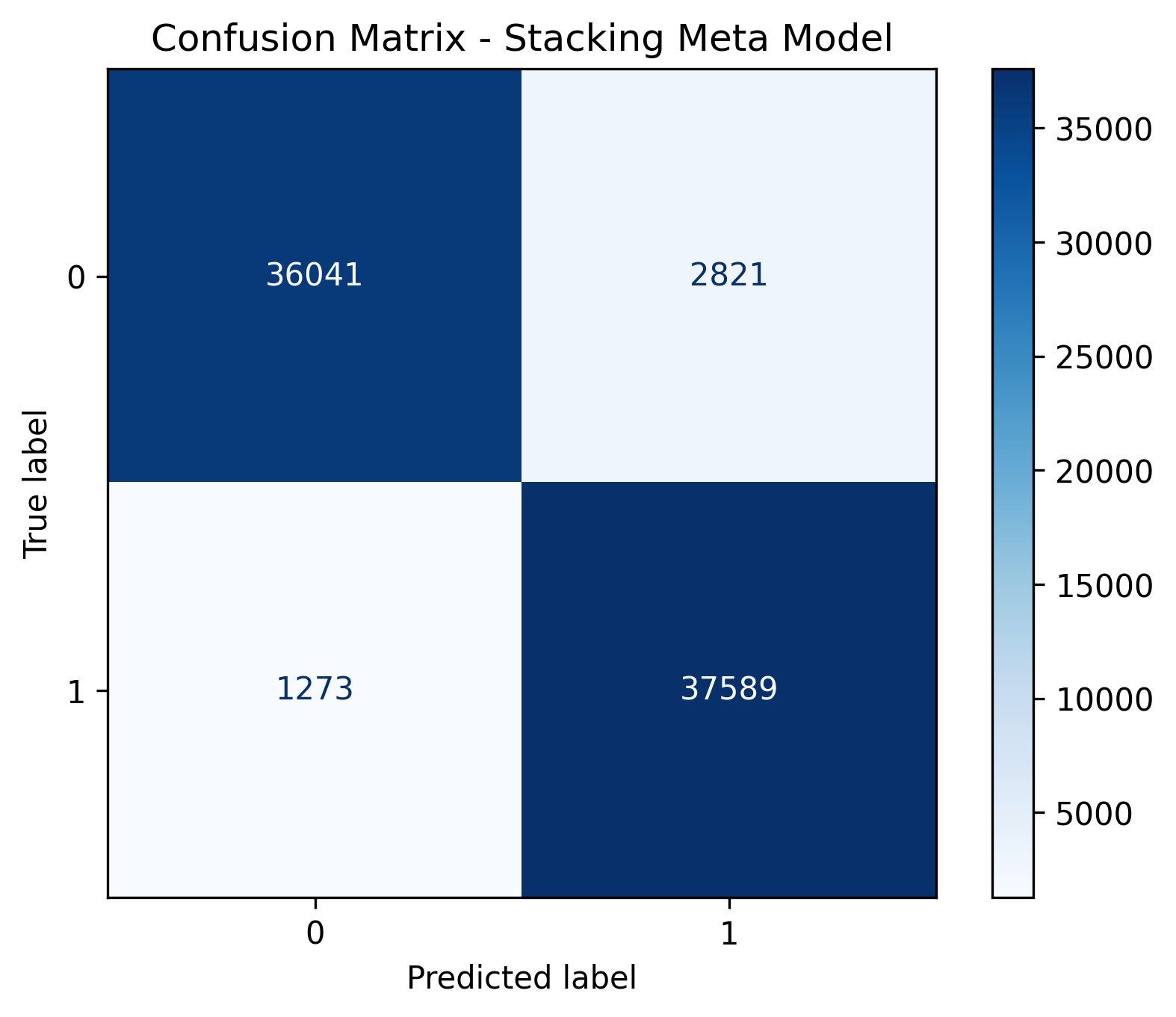}
\caption{Performance of the Stacking Model: Confusion Matrix}
\label{fig:cm}
\end{figure}

Figure~\ref{fig:cm}, shows the stacking model achieved 36,041 true negatives and 37,589
true positives, with few misclassifications: 2,821 false positives and 1,273
-neg), while performing with balance on both.
classes.

\begin{figure}[h!]
\centering
\includegraphics[width=0.6\textwidth]{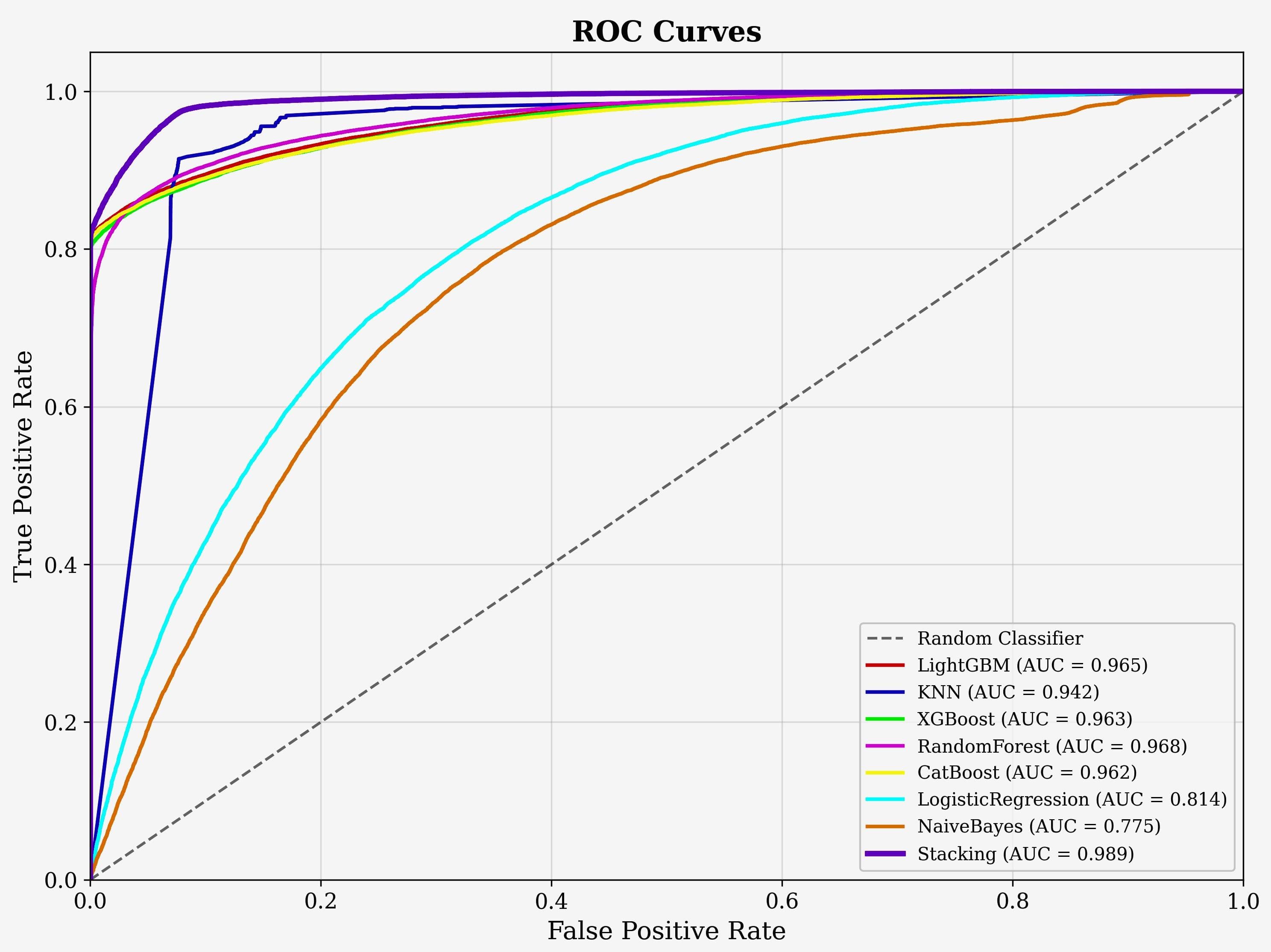}
\caption{Comparison of ROC Curves for All Models}
\label{fig:rocC}
\end{figure}

ROC curve Figure~\ref{fig:rocC}, shows the capability of each model in differentiating between diabetic and non-diabetic patients. The stacking ensemble dominates with
an AUC of 0.989 - i.e., it separates the two classes nearly perfectly.
across all decision thresholds. Random Forest (AUC = 0.968) and LightGBM
AUC = 0.965), while traditional methods such as Naive Bayes also perform well.
While LDA, SVM, and Logistic Regression are far behind at AUCs of 0.775 and 0.814, respectively.

\begin{figure}[h!]
\centering
\includegraphics[width=0.6\textwidth]{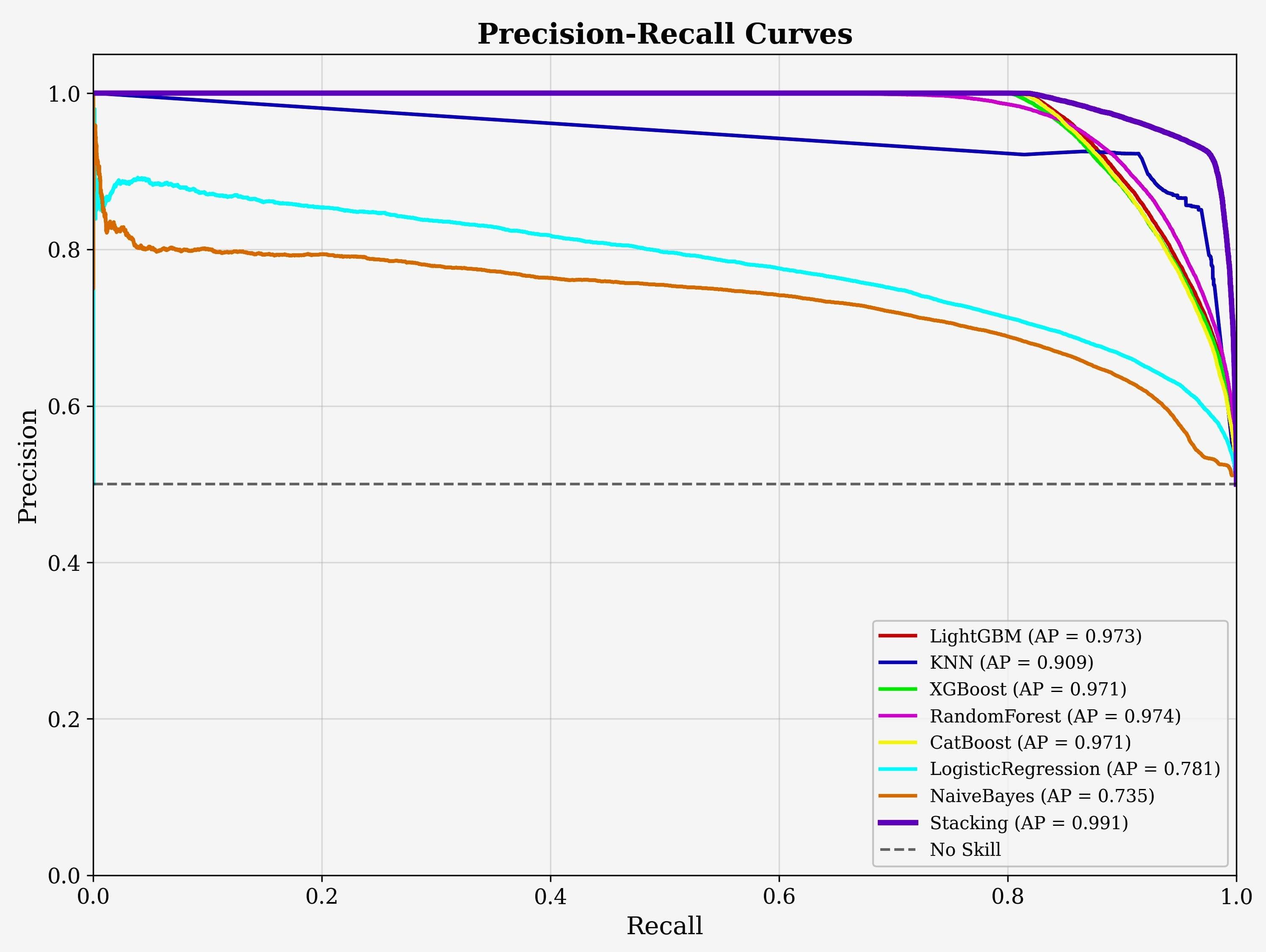}
\caption{Comparison of Precision–Recall Curves for All Models}
\label{fig:pr4}
\end{figure}

The clear separation reflects why ensemble methods are superior for the clinical task at hand. Precision-Recall Curve Figure~\ref{fig:pr4} tells the real-world story.
for medical applications which involve catching diabetic cases, recall while minimizing
Both false alarms and precision are important. The stacking model realizes an
an average precision of 0.991, which is the best for all models. This excellent performance indicates that the stacking ensemble reliably identifies diabetic patients while
high confidence for the positive predictions it makes. Random Forest (AP =
0.974) and LightGBM (AP = 0.973) follow closely, but the stacking approach’s
0.991 score confirms its superior reliability for clinical deployment.

\subsection{Comparison with Existing Works}

Our proposed method outperforms previous work in terms of dataset size, hyperparameter tuning, feature selection, and imbalance handling. While some papers proposed slightly higher accuracy in different datasets with small numbers of data, our proposed model is highly competitive with large datasets (253,680 patients). Two papers proposed somewhat worse performance than our model: they achieved 86.6\% and 92.5\% accuracy, while we achieved 94.82\% accuracy with a better AUC-ROC (0.9895). Table~3 contains a comprehensive summary of prior work.

\begin{table}[ht]
\centering
\caption{Performance comparison of proposed stacking model with previous studies in terms of dataset size, dataset source, and accuracy.}
\begin{tabular}{lccc}
\toprule
\textbf{Reference} & \textbf{Dataset} & \textbf{Method} & \textbf{Accuracy (\%)} \\
\midrule
Rahaman \cite{6} & 768 (PIMA) & Random Forest & 81 \\
Gupta \cite{7} & 768 (PIMA) & Logistic Regression & 86 \\
Ha \cite{8} & PIMA + hospital + wearables & Random Forest & 90 \\
El Massari \cite{9} & 768 (PIMA) & Ontology Classifier & 77.5 \\
Khokhar \cite{10} & 253,680 (BRFSS) & Ensemble model & 92.5 \\
Ren \cite{11} & 253,680 (BRFSS) & CatBoost Classifier & 86.6 \\
\textbf{Our Model} & \textbf{253,680 (BRFSS)} & \textbf{KNN + XGBoost } & \textbf{94.82} \\
\bottomrule
\end{tabular}
\label{tab:merged_comparison_dataset}
\end{table}

In comparison to previous studies, our model achieves a good balance of high prediction performance and scalability.

\subsection{Deployment Interface}

\begin{figure}[ht]
\centering
\begin{minipage}{0.3\textwidth}
  \centering
  \includegraphics[height=5cm,width=\linewidth,keepaspectratio]{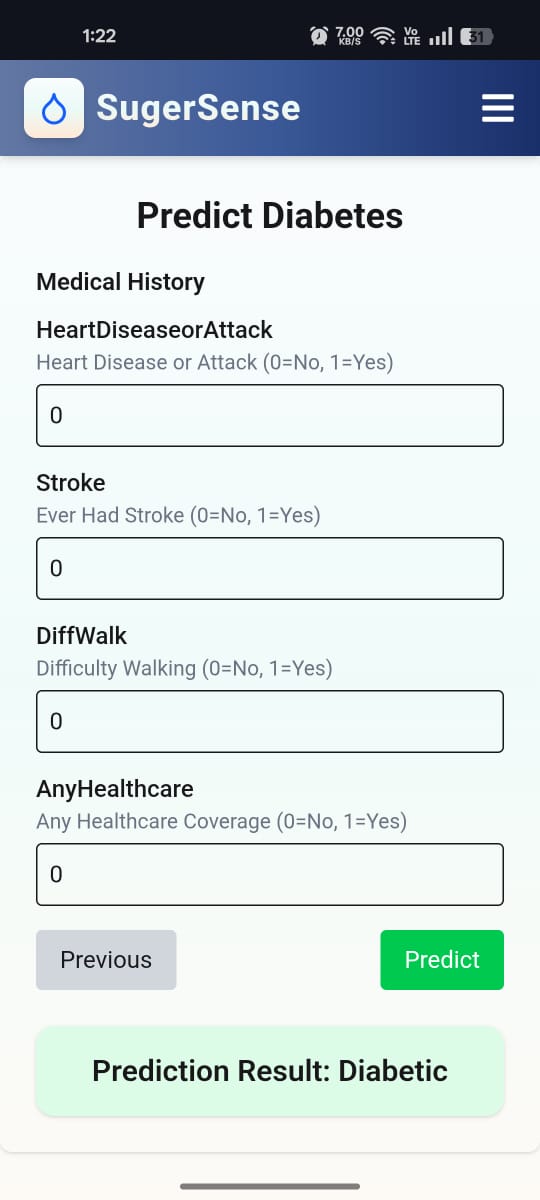}
  \caption{User-provided clinical factors and model prediction results.}

\end{minipage}\hfill
\begin{minipage}{0.3\textwidth}
  \centering
  \includegraphics[height=5cm,width=\linewidth,keepaspectratio]{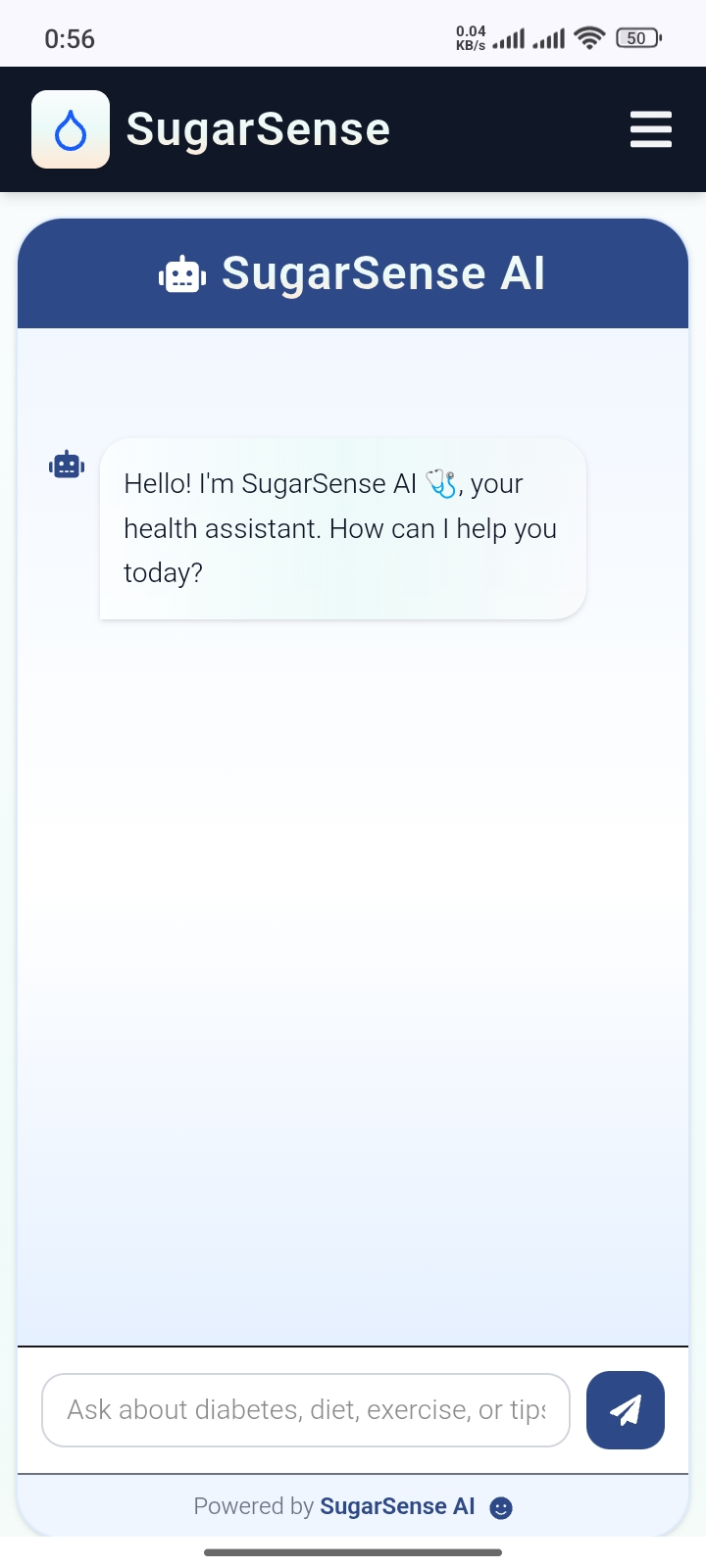}
  \caption{\textit{SugarSense} chatbot providing real-time health assistance.}
 
\end{minipage}\hfill
\begin{minipage}{0.3\textwidth}
  \centering
  \includegraphics[height=5cm,width=\linewidth,keepaspectratio]{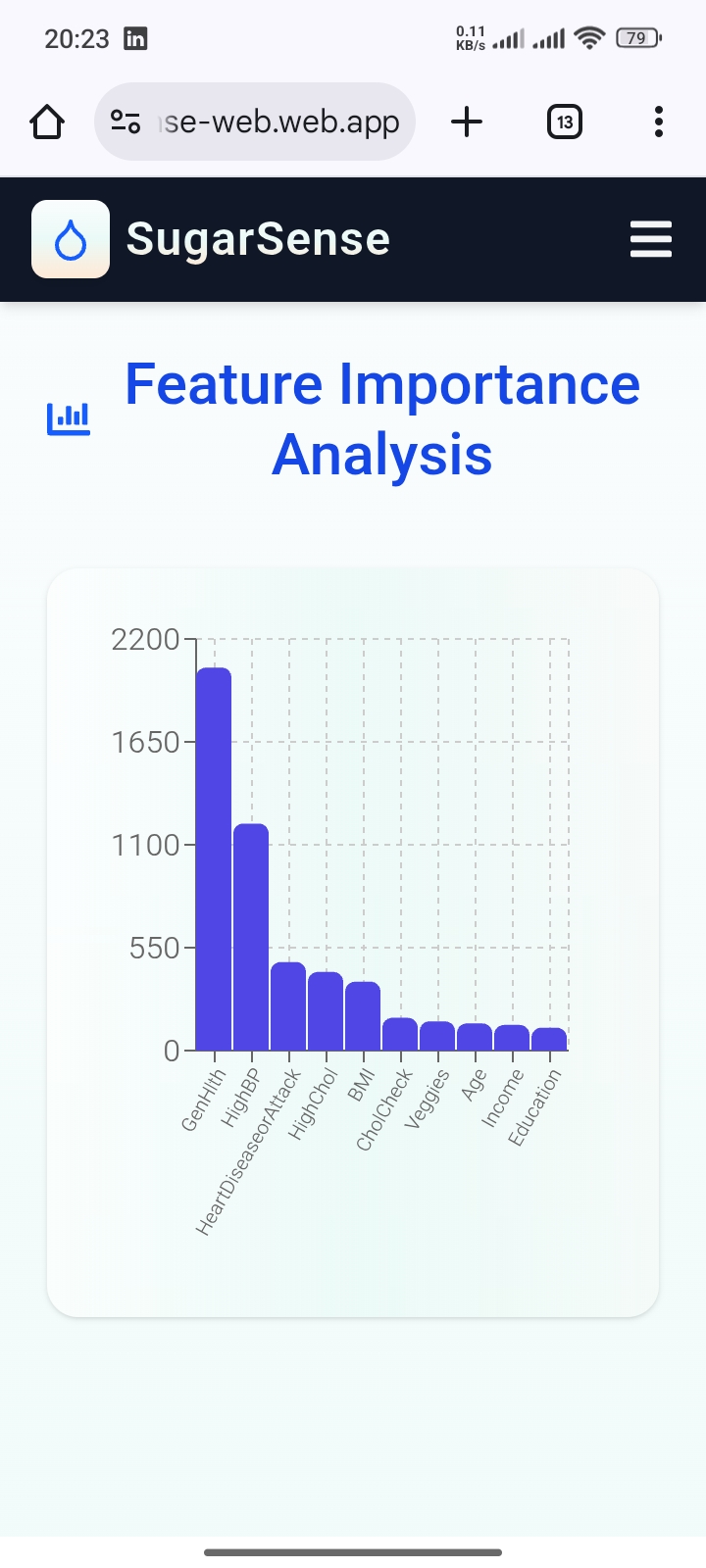}
  \caption{Feature importance visualization for key risk factors.}

\end{minipage}
\end{figure}

In this study, a user-friendly web-based application was developed to demonstrate the practical implementation of the proposed diabetes prediction model. The application allows users to input personal and health-related data such as age, gender, BMI, blood pressure, glucose level, and insulin measurements. After submission, the backend machine learning model analyzes the input data in real time and predicts whether the individual is diabetic or non-diabetic. As illustrated in Figure 5,6, and 7, the web interface displays the prediction outcome along with a confidence score, ensuring result transparency and interpretability.

\section{Discussion}
In this study, our model has achieved the highest accuracy of  94.82\%, which is better than previous works. Moreover, we achieved an ROC-AUC of 0.98 and a PR-AUC of  0.991. For handling the imbalanced data, we used SMOTE and Tomek Links. We implemented multiple supervised learning algorithms, including logistic regression, random forest, gradient boosting, XGBoost, LightGBM, CatBoost, KNN, and SVM, in developing predictive models for diabetes, and among the tested algorithms, the ensemble with XGBoost and KNN provided the best output. In conclusion, we developed a web-based application that will help healthcare professionals with diabetes prediction.   

The BRFSS dataset is a survey-based dataset, which has reporting bias and missing entries. The data is highly imbalanced. Although we used SMOTE and Tomek Links to improve the dataset, the oversampling may not be perfectly reflective.  This database, from 2015, is based on only one population. Performance may vary for different communities. The combination of datasets must be reliable. In the future, we can use cost-sensitive learning or adaptive boosting for imbalanced data instead of relying solely on SMOTE and Tomek Links and test this model on different populations or ethnic groups.

\section{Conclusion}
This study demonstrates the prediction of diabetes using the 2015 BRFSS dataset, which contains 253,680 records with 22 numerical features. SMOTE and Tomek Links were used to rectify the imbalanced data. We explored supervised learning algorithms, including logistic regression, random forest, gradient boosting, XGBoost, LightGBM, CatBoost, KNN, and SVM, in developing predictive models for diabetes, and among the tested algorithms, the ensemble with XGBoost and KNN provided the best accuracy of 94.82\%, ROC-AUC of 0.989, and PR-AUC of 0.991. We developed an Web-based application. 

This study shows that using hybrid feature selection with balanced data improves the diabetes classification performance. The data is very imbalanced. Although we used SMOTE and Tomek Links to improve the dataset, the oversampling may not be perfectly reflective. The BRFSS dataset is a survey-based dataset, which has reporting bias and missing entries. This database, from 2015, is based on a population. Performance may vary for different populations or ethnic groups. The combination of datasets must be reliable. For ensuring global applicability, the model may be tested on other countries and different ethnic groups, and may use cost-sensitive learning or adaptive boosting for imbalanced data instead of relying solely on SMOTE and Tomek Links.

\end{document}